\documentclass[10pt,twocolumn,letterpaper]{article}
\usepackage{amsmath}
\usepackage{bm}

\usepackage[accsupp]{axessibility} 
\usepackage{adjustbox}
\usepackage{graphicx}
\usepackage{amsmath}
\usepackage{amssymb}
\usepackage{booktabs}
\usepackage[pagenumbers]{cvpr} 
\usepackage{indentfirst} 
\usepackage[dvipsnames]{xcolor}

\definecolor{cvprblue}{rgb}{0.21,0.49,0.74}
\usepackage[pagebackref,breaklinks,colorlinks,citecolor=cvprblue]{hyperref}
\usepackage{natbib}
\setcitestyle{numbers,square}

\def\paperID{5279} 
\def\confName{CVPR}
\def\confYear{2024}

\title{LPSNet: End-to-End Human Pose and Shape Estimation with Lensless Imaging}

\author{Haoyang Ge$^{1,\dagger}$, Qiao Feng$^{1,\dagger}$, Hailong Jia$^{1}$, Xiongzheng Li$^{1}$, Xiangjun Yin$^{1}$,\\ You Zhou$^{2}$, Jingyu Yang$^{1}$, Kun Li$^{1,*}$ \\
$^{1}$Tianjin University, China   \quad  $^{2}$Nanjing University, China\\
{\tt\small$\left\{\right.$ghy0623,fengqiao,jhl,lxz,yinxiangjun,yjy,lik$\left\}\right.$@tju.edu.cn} \quad{\tt\small zhouyou@nju.edu.cn}
}

\begin{document}
\twocolumn[{
\renewcommand\twocolumn[1][]{#1}
\maketitle
\begin{center}
    \captionsetup{type=figure}
    \includegraphics[width=1\textwidth]{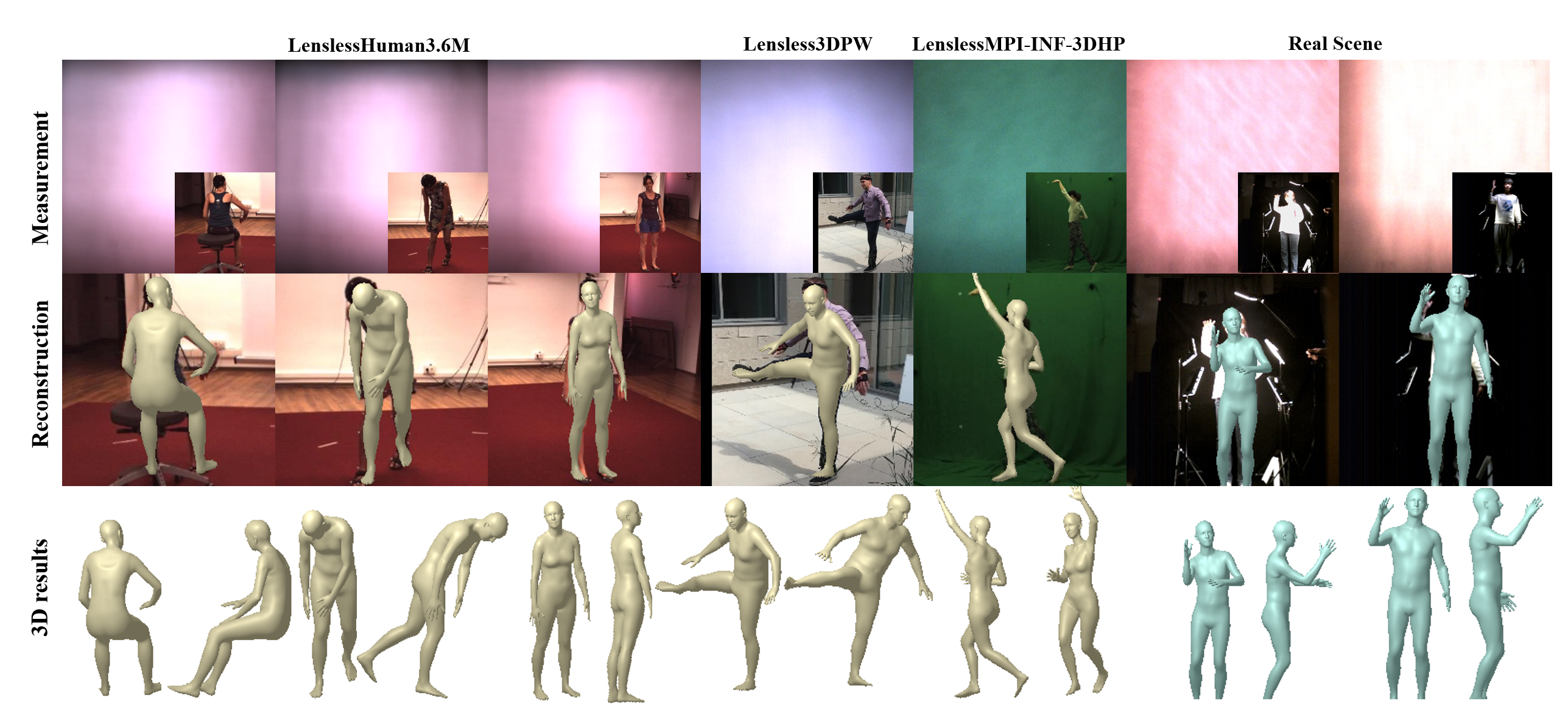}
    \captionof{figure}{\textbf{LPSNet: End-to-End Human Pose and Shape Estimation with Lensless Imaging.} We contribute a framework for estimating human poses and shapes from individual lensless measurements. The first row shows the input measurements acquired by our lensless imaging system, the second row shows the estimated human poses and shapes from lensless measurements, and the bottom row shows the 3D results shown in different views.}
    \label{head}
\end{center}
}]


\makeatletter\def\Hy@Warning#1{}\makeatother
\let\thefootnote\relax\footnotetext{$\dagger$ Equal contribution.}

\let\thefootnote\relax\footnotetext{* Corresponding author.}

\begin{abstract}
Human pose and shape (HPS) estimation with lensless imaging is not only beneficial to privacy protection but also can be used in covert surveillance scenarios due to the small size and simple structure of this device.
However, this task presents significant challenges due to the inherent ambiguity of the captured measurements and lacks effective methods for directly estimating human pose and shape from lensless data.
In this paper, we propose the first end-to-end framework to recover 3D human poses and shapes from lensless measurements to our knowledge.
We specifically design a multi-scale lensless feature decoder to decode the lensless measurements through the optically encoded mask for efficient feature extraction.
We also propose a double-head auxiliary supervision mechanism to improve the estimation accuracy of human limb ends.
Besides, we establish a lensless imaging system and verify the effectiveness of our method on various datasets acquired by our lensless imaging system. The code and dataset are available at \href{https://cic.tju.edu.cn/faculty/likun/projects/LPSNet}{https://cic.tju.edu.cn/faculty/likun/projects/LPSNet}.
\end{abstract}    
\section{Introduction}
\label{sec:intro}

In recent years, lensless imaging technologies~\cite{ozcan2016lensless,boominathan2016lensless,boominathan2020phlatcam,kuo2017diffusercam} have advanced significantly due to their many advantages, such as privacy protection, smaller size, simple structure, and lower cost.
3D human pose and shape (HPS) estimation~\cite{zhang2020learning,zhang2021pymaf,Goel_2023_ICCV,kanazawaHMR18} requires miniaturized and lightweight imaging system as application scenarios become more diverse.
All these advantages, especially privacy, make the lensless imaging system very suitable as an imaging device for human pose and shape estimation.
In this work, we propose LPSNet, which aims to estimate 3D human pose and shape from lensless measurements instead of RGB images, achieving cheaper and privacy-protecting 3D human pose and shape estimation.

A thin, lightweight, and potentially cost-effective optical encoder is used in a lensless imaging system instead of traditional cameras with lenses, while others are expensive, rigid, and occupy more space.
At this stage, the application of a lensless imaging system is extensive, it is mainly used in the field of microscopic imaging, RGB image reconstruction, and so on.
More valuable information can be obtained from lensless measurements due to the special optical encoding method of lensless imaging systems.
Directly estimating human pose and shape from lensless measurements is not currently possible.
The first step is to reconstruct an RGB image from a lensless measurement and then estimate the human pose and shape from the RGB images.
However, experimental findings indicate that the reconstructed RGB images are of suboptimal quality, resulting in incomplete local features and significant deviations in the position of the human body.
Combining these factors leads to inaccurate human pose estimation when using lensless measurements to reconstruct RGB images.
This approach has computational burden and computational resources, making it very unfavorable for deployment at the endpoint.

In this paper, We aim to advance human pose and shape estimation using a lensless imaging system, which needs to overcome two main challenges. First, how to extract features from lensless measurements for human pose and shape estimation. 
Secondly, during early experiments, when using features extracted from lensless measurements to estimate human pose and shape, we found poor estimation accuracy of human limbs.

To address these challenges, we introduce LPSNet, the first end-to-end human pose and shape estimation framework with lensless imaging.
To extract features from optically encoded lensless measurements, we propose a multi-scale lensless feature decoder (MSFDecoder).
Specifically, we introduce a global perception layer to enhance the global decoding capability of MSFDecoder.
The global information that has been optically encoded to the global can be efficiently decoded to obtain a feature map that can be used in subsequent processes.
To improve pose and shape estimation, we propose a Double-Head Auxiliary Supervision(DHAS) mechanism to be implemented during training.
Auxiliary supervision can improve the estimation accuracy of human limbs and correct results with large deviations.

Our main contributions can be summarized as follows:
\begin{itemize}
  \item [1)] 
  We propose LPSNet, an end-to-end pose and shape estimation network for lensless imaging systems.
  This is the first work to estimate human poses and shapes directly from lensless measurements.
  
  \item [2)]
  We propose MSFDecoder, a Multi-Scale Lensless Feature Decoder that decodes and extracts features from lensless measurements, which can be receptive to global features in lensless measurements.
  \item [3)]

  We propose a Double-Head Auxiliary Supervision mechanism for both pose and shape estimation, which can improve the estimation accuracy of human limbs.

\end{itemize}

\section{Related Work}

\subsection{Lensless Imaging System}
A conventional photographic camera typically comprises a focusing lens, which may consist of one or more optical elements and an image sensor positioned at or near the focal length of the lens. The lens in such a camera directs the projected light from the observed scene onto the sensor, aiming to accurately map specific scene points to individual pixels on the sensor.
Conversely, in a lensless imaging system, the absence of a lens defines its configuration. Instead, an optical modulator, such as a coded amplitude mask or a diffuser, is positioned between the scene and the image sensor, often close to the sensor itself. As a result, the recorded data deviates significantly from the expected RGB image during imaging. In this process, the local information of the object transforms overlapping global information through the optically encoded mask.

Within the mask-modulated lensless systems, a fixed optical mask is introduced to create a versatile lensless system that can work for a wide range of object distances and lighting scenarios, whether passive or uncontrolled.
The mask modulates the incoming light and generates a measurement that can be decoded through computational methodologies.
Mask-modulated lensless imaging systems were used to perform 2D imaging~\cite{asif2016flatcam,boominathan2020phlatcam,kuo2017diffusercam}, refocusing~\cite{boominathan2020phlatcam}, 3D imaging~\cite{boominathan2020phlatcam,antipa2018diffusercam}, and microscopic imaging~\cite{boominathan2020phlatcam,antipa2018diffusercam,adams2017single}.

\begin{figure}[t]
  \centering
   \includegraphics[width=0.98\linewidth]{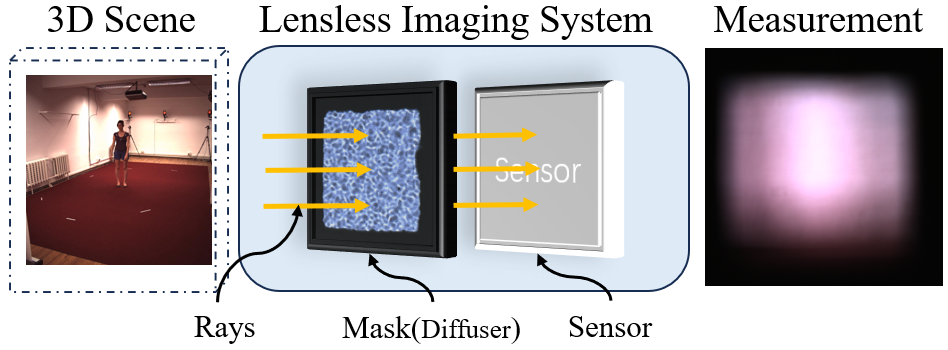}

   \caption{The workflow of the lensless imaging system and the final measurement obtained is the result obtained from the encoding of 3D scene information by the lensless imaging system.
   The optically encoded mask transforms local information in the 3D scene into overlapping global information.}
   \label{fig:3.2.1}
\end{figure}

Generally speaking, amplitude modulators and phase modulators~\cite{asif2016flatcam,adams2017single} are the two types of optical masks used in lensless imaging systems.
Phase modulators can be further sub-categorized into phase gratings~\cite{stork2013lensless,stork2014optical}, diffusers~\cite{antipa2018diffusercam}, and phase masks~\cite{boominathan2020phlatcam}. One key characteristic of a mask-modulated lensless system is the pattern the mask produces on the sensor for a point light source in the scene. We call this pattern the Point-Spread Function(PSF), and its properties determine the imaging model of the system.
As shown in ~\cref{fig:3.2.1}, we design a simple mask-modulated lensless system for data acquisition in this experiment.
The lensless imaging system designed in our experiments chose a diffuser as the mask.

The main advantages of lensless imaging are as follows: lensless is small in size and can be assembled in a variety of miniaturized equipment; lensless is light and easy to carry; lensless cameras are still cheap to make; lensless cameras have a wide view field, larger than traditional wide-angle cameras; It can also protect the user's privacy to a great extent.

\subsection{Human Pose and Shape Recovery}
There are two primary methods for human pose and shape estimation: optimization-based approaches and regression-based approaches.
Numerous methodologies have been developed to estimate the three-dimensional human pose and shape by employing iterative optimization techniques, \textsl{e.g.}, SMPLify~\cite{Bogo:ECCV:2016} and variations of the SMPLify~\cite{Lassner:UP:2017,rempe2021humor,tiwari22posendf,SMPL-X:2019}.

Many existing Regression-based methods follow the architecture of HMR~\cite{kanazawaHMR18}, which uses a pre-trained backbone to extract image features followed by regression to obtain SMPL~\cite{Bogo:ECCV:2016} parameters.
Many improvements to the original method have been proposed since its introduction.
In particular, many papers have proposed alternative methods for pseudo-ground truth generation, including using temporal information~\cite{arnab2019exploiting}, multiple views~\cite{leroy2020smply}, or iterative optimization~\cite{kolotouros2019learning,joo2021exemplar,pavlakos2022human}. SPIN~\cite{kolotouros2019learning} proposed an in-the-loop optimization that incorporated SMPLify~\cite{Bogo:ECCV:2016} in the HMR training, which combines the optimization-based approach and the regression-based approach for human pose and shape estimation.
PyMAF~\cite{zhang2021pymaf,zhang2019predicting} proposes a pyramidal mesh alignment feedback loop for regression of SMPL parameters.
Our approach references the design of PyMAF in the part of the SMPL parameters regression.
PARE~\cite{kocabas2021pare} proposes a body-part-guided attention mechanism for better occlusion handling.
HKMR~\cite{georgakis2020hierarchical} performs a prediction guided by the hierarchical structure of SMPL.
HMR2.0~\cite{Goel_2023_ICCV} employs a large training dataset and proposes a fully transformer-based approach for 3D human pose and shape estimation from a single image.
Many related approaches make non-parametric predictions, i.e., instead of estimating the parameters of the SMPL model, they explicitly regress the vertices of the mesh. GraphCMR~\cite{kolotouros2019convolutional} uses a graph neural network for the prediction, METRO~\cite{lin2021end} and FastMETRO~\cite{cho2022cross} use a transformer, while Mesh Graphormer ~\cite{lin2021mesh} adopts a hybrid between the two. 

In this paper, we propose the first work to perform end-to-end human pose and shape estimation from lensless measurements.
We designed a multi-scale lensless feature decoder to decode the lensless measurements based on a mask-encoder to obtain more efficient features.
We also propose a double auxiliary supervision mechanism to improve the estimation accuracy of human limbs.

\section{Method}

\subsection{Overview}
\begin{figure*}[t]
  \centering
   \includegraphics[width=0.95\linewidth]{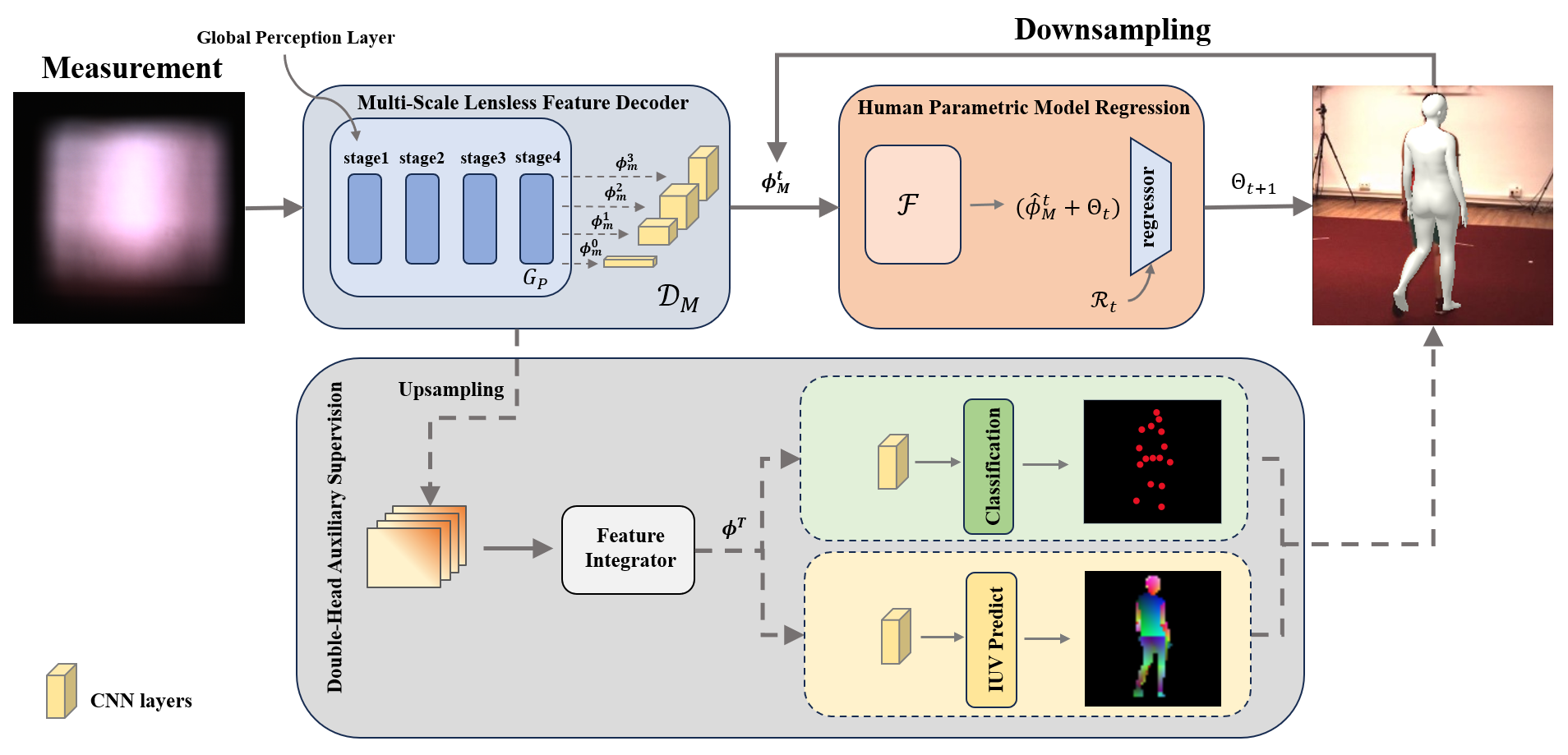}
   \caption{\textbf{Overview of the proposed framework.}
   A measurement $M$ is passed through a Multi-Scale Lensless Feature Decoder to get spatial characteristics at different scales.
   These feature maps are fed into the regressor for human pose and shape estimation.
   Also, these feature maps are fed into the Double-Head Auxiliary Supervision to improve the estimation accuracy.
   }
   \label{fig:3.1.1}
\end{figure*}

Our work focuses on human pose and shape estimation from lensless measurements. In this section, we present the technical details of our approach.
As depicted in \cref{fig:3.1.1}, the core of our method involves the following three components:
1) a Multi-Scale Lensless Feature Decoder(MSFDecoder) (Sec. \ref{3.2}) that can effectively decode the information encoded by the lensless imaging system;
2) a human parametric model regressor (Sec. \ref{3.3}) that takes the multi-scale features produced by MSFDecoder as input and predicts the SMPL parameters;
3) a Double-Head Auxiliary Supervision mechanism(DHAS) (Sec. \ref{3.4}) that can assist the LPSNet to improve the estimation accuracy of human limbs.

\subsection{Multi-Scale Lensless Feature Decoder}
\label{3.2}
As illustrated in ~\cref{fig:3.2.1}, the optical information of the subject is diffused by the mask and projected onto the sensor. The process of obtaining measurements through the lensless imaging system can be treated as encoding optical information in the scene into lensless measurements.
The goal of our lensless feature decoder $D_M$ is to decode these lensless measurements to multi-scale features, which are then utilized for subsequent human body prediction.
The design of the global perception layer $G_P$ is inspired by HRNet~\cite{sun2019deep}.
The global perception layer inherits many of the advantages of HRNet~\cite{sun2019deep} and can always maintain a high resolution, while the information interaction between different branches complements the information loss caused by the reduction in the number of channels.
These advantages are important for extracting features from lensless measurements.
Then, we design different convolutional layers for different scales of features that are used to refine the features.
 
Formally, the decoder takes a lensless measurement $M$ as input and decodes the information encoded by the lensless imaging system. 
The global perception layer ensures the integrity of global information for each pixel by maintaining the high-resolution convolution branch and the low-resolution convolution branch in parallel. It further enhances the information extraction accuracy by repeating the multi-scale fusion across parallel convolutions.
As shown in ~\cref{fig:3.1.1}, the global perception layer takes $M$ as input and generates a set of multi-scale spatial features $\bm{\phi}^{t = 0,1,2,3}_m$ as input for the subsequent steps.
The final output $\bm{\phi}^t_M$ of the decoder is obtained by passing $\bm{\phi}^t_m$ into different convolutional layers,
\begin{equation}
\phi^{t}_M = D_M\left({M} \right).
\end{equation}
With different scales of decoding, we can obtain feature maps at different granularities, through which we can perceive the location of people in the scene.
The results obtained by the lensless decoder are utilized for subsequent predictions with the SMPL model, incorporating pose, shape, and camera parameters $\Theta=\{\boldsymbol{\theta}, \boldsymbol{\beta}, \boldsymbol{\widetilde{\pi}\}}$, where $\boldsymbol{\widetilde{\pi}}$ is pseudo-camera parameters for the subsequent projection calculation.

\subsection{Human Parametric Model Regression}
\label{3.3}
Our human parametric model regression is mainly inspired by the improved Human Mesh Regression described in PyMAF~\cite{zhang2021pymaf}.
Different from the pyramid features obtained by deconvolution in PyMAF~\cite{zhang2021pymaf}, our global perception layer exploits HRNet~\cite{kanazawaHMR18} to maintain more high-resolution information.
We use a set of sampling points $S_t$ and project them onto the feature maps to extract point-wise features.
The point-wise features corresponding to each point $s$ in $S_t$ are bilinearly sampled from $\bm{\phi}^{t = 1,2,3}_M$.
(Note that $\bm{\phi}^{t = 0}_M$ is unnecessary here and is used as input for subsequent initialization.)
Then the dimensions of these features will be reduced by MLP(multi-layer perceptron) and concatenated as a feature vector $\bm{\hat{\phi}}^t_M$ which can be formulated as:
\begin{equation}
\begin{split}
\hat{\phi}^t_M &=\mathcal{F}\left(\phi^t_M, S_t\right) \\ 
&=\oplus\left\{f\left(\phi_M^t(s)\right), \text { for } s \text { in } S_t\right\},
\end{split}
\end{equation}
where $\mathcal{F}(\cdot)$ denotes the overall point-wise feature extraction, $f(\cdot)$ is the MLP layer, and $\oplus$ denotes the concatenation.
We use an iterative mechanism to complete the interaction of different scale features.
Sequentially, a parameter regressor $\mathcal{R}_t$ takes features $\hat{\phi}^t_M$ and the current parameter estimation $\Theta_{t}$ as inputs and outputs the parameter residual. Parameters are updated as $\Theta_{t+1}$ by adding the residual to $\Theta_{t}$ which can be formulated as:
\begin{equation}
\begin{split}
\Theta_{t+1} =\Theta_{t} + \mathcal{R}_t\left(\Theta_{t} , \hat{\phi}^t_M \right), \text { for } t\, > \,0.
\end{split}
\end{equation}
The initial parameter $\Theta_{0}$ is initialized by feeding $\bm{\phi}^{t = 0}_M$ into the regressor $\mathcal{R}_t$.

When the predicted parameters $\Theta$ (the subscript \textit{t} is omitted for simplicity) are obtained through each scale feature, a mesh with vertices of $V=\mathcal{M}(\boldsymbol{\theta}, \boldsymbol{\beta}) \in \mathbb{R}^{N \times 3}$ can be generated accordingly, where $N = 6890$ is the number of vertices in the SMPL model.
We downsample $V$ to get the sampling point $S_{t+1}$ for the next iteration.
These mesh vertices can be mapped by pre-trained linear regressor to obtain sparse 3D keypoints $J \in \mathbb{R}^{N_j \times 3}$.
With the estimated pseudo-camera parameters, we can obtain 2D keypoints $K \in \mathbb{R}^{N_j \times 2}$ by projecting $J$ on the measurement coordinate system.
We calculate the loss between the 2D keypoints and the ground truth, in this way reducing the difference between the 2D projection and the human pose in the real scene.
Concurrently, supplementary 3D supervisions regarding 3D keypoints and model parameters are integrated when authentic 3D labels are available. The total loss function for the parameter regressor is thereby formulated as follows:

\begin{equation}
\mathcal{L}_{\text {reg }}=\lambda_{2 d}\|K-\hat{K}\|_{2}+\lambda_{3 d}\|J-\hat{J}\|_2+\lambda_{\text {para }}\|\Theta-\hat{\Theta}\|_2,
\end{equation}
where $\|\cdot\|_2$ is the squared L2 norm, $\hat{K}$,$\hat{J}$, and $\hat{\Theta}$ denote the ground truth 2D keypoints, 3D keypoints, and model parameters, respectively. 

\subsection{Double-Head Auxiliary Supervision}
\label{3.4}
The spatial feature map of the human body is relatively rough and contains a lot of noise. Therefore, there are still some deviations in the perception of human limbs and pose. To improve our estimation accuracy of human limbs, we introduce a Double-Head Auxiliary Supervision (DHAS) mechanism to obtain finer spatial features during the training stage.

Specifically, we first convert all the spatial features at different scales to a single feature map $\phi^T$ by upsampling and then concatenating them together. The feature map $\phi^T$ is used for the different auxiliary supervision heads. 
On the one hand, we generate a heat map representation by a classification layer to explicitly indicate the position of the 2d keypoints.
On the other hand, we also estimate a dense map through the IUV Predict layer to get a dense correspondence between the SMPL model and the feature map.
The loss function for the Double-Head Auxiliary Supervision consists of two parts, which can be written as:
\begin{equation}
\mathcal{L}_{d a s}  =\mathcal{L}_{s c} + \mathcal{L}_{den}.
\end{equation}

\textbf{Keypoints Supervision.}
We utilize the SimCC-based algorithm~\cite{li2022simcc} for predicting pose keypoints. This approach treats keypoint localization as a classification task in horizontal and vertical coordinates. During training, instead of estimating the actual coordinates, we employ two vectors, \textit{x} and \textit{y}, and convert the ground truth 2D keypoints into such vectors to compute the loss.
The loss function is then formulated as follows:
   \begin{equation}
    \begin{aligned}
    \mathcal{L}_{sc} =\lambda_{x y} \left(\operatorname{KL-Loss}({x} , \hat{x}) + \operatorname{KL-Loss}({y} , \hat{y})\right),
    \end{aligned}
    \end{equation}
where KL-Loss is the Kullback-Leibler divergence loss, $\hat{x}$ and $\hat{y}$ denote the processed ground truth 2D keypoints, respectively.

\textbf{IUV Supervision.}
We adopt the IUV mapping defined in DensePose~\cite{guler2018densepose} as the dense correspondence representation. This mapping establishes deterministic correspondences between foreground pixels in 2D images and vertices on 3D surfaces. The vertices on the template mesh can be mapped back to pixels on the foreground using a predefined bijection mapping between 3D surface space and 2D UV space.
The dense correspondence representation comprises the index of body parts $P$ and UV values of mesh vertices. During training, we apply classification loss to the index of body parts $P$ and regression loss to the UV channels of the dense correspondence mapping.
The loss function is formulated as follows:
  \begin{equation}
    \begin{aligned}
    \mathcal{L}_{den} & =\lambda_{p i} \text { CrossEntropy }(P, \hat{P}) \\
    & +\lambda_{u v} \operatorname{SmoothL1}( U,  \hat{U}) \\
    & +\lambda_{u v} \operatorname{SmoothL1}( V, \hat{V}) \\
    \end{aligned}
    \end{equation}
where $\hat{P}$,$\hat{U}$, and $\hat{V}$ denote the ground truth of $P$ and $UV$ values, respectively.

\section{Experiments}

\begin{figure}[t]
  \centering
   \includegraphics[width=0.9\linewidth]{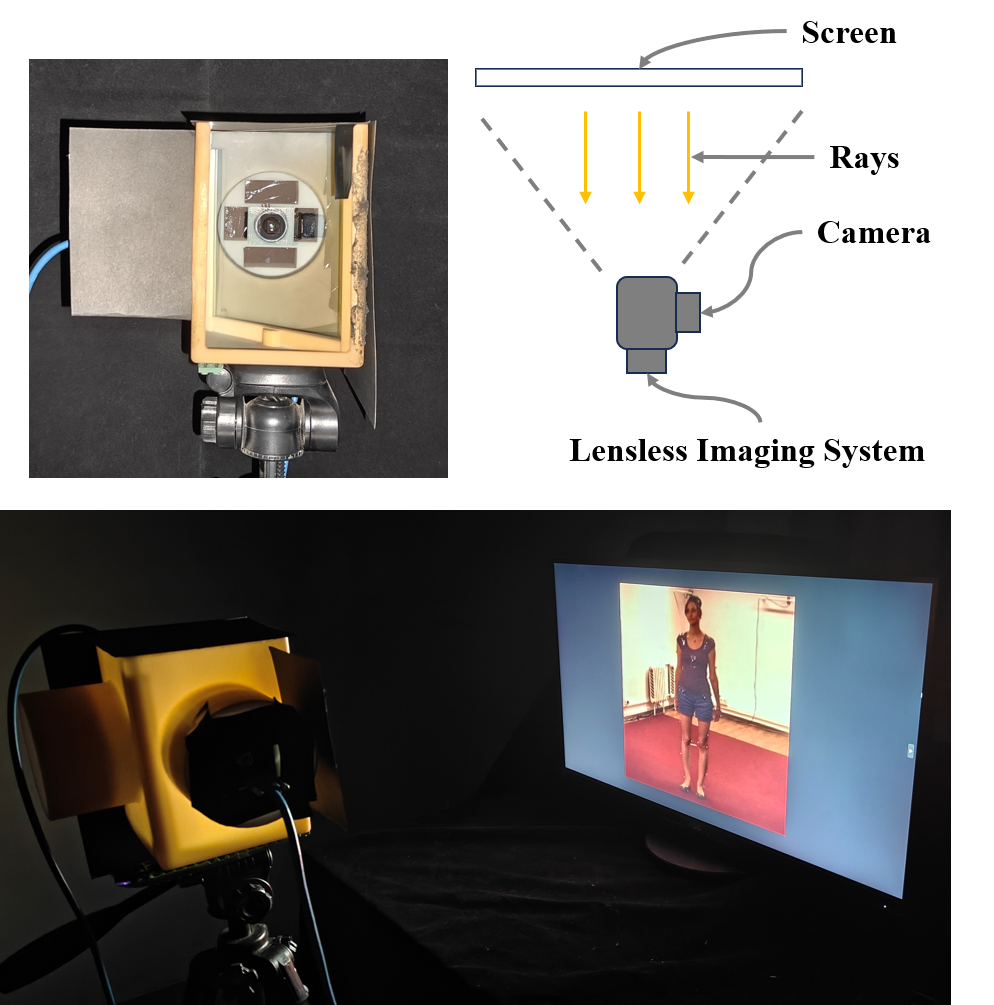}

   \caption{
   On the top left is a figure of the dual camera system we created. This system consists of an RGB camera and a lensless imaging system using a diffuser as a mask. The top right shows the process of collecting our dataset. The image at the bottom is a record of the process of collecting our dataset.}
   \label{fig:4.3}
\end{figure}

\subsection{Imaging System and Dataset}
\textbf{Lensless Imaging System.} 
Due to the scarcity of human pose and shape datasets based on lensless imaging systems, we developed a basic lensless system to gather experimental data. 
The system, illustrated in ~\cref{fig:4.3}, comprises two primary components: a lensless imaging system for capturing lensless measurements and an RGB camera for capturing real images corresponding to the lensless measurements. To ensure identical light paths, we employed a beam splitter between the lensless sensor and the RGB camera. This allowed for consistent imaging conditions and facilitated accurate correlation between the lensless measurements and the corresponding real images.

For the lensless imaging part of the system, we adopted the Mask-Modulated Lensless System, inspired by design options such as diffusercam~\cite{antipa2018diffusercam} and phlatcam~\cite{boominathan2020phlatcam}. Specifically, we selected a diffuser as our mask and incorporated PhlatCam's concept of using larger sensors to enhance global information perception.

\textbf{Datasets.}
The inputs to LPSNet are lensless measurements. However, datasets for classical human pose estimation are not directly available at present. Leveraging the lensless imaging system we have constructed, and the mathematical model of the system's imaging, the sources of datasets for our experiments can be categorized as follows:
\begin{itemize}   
  \item 
  \textbf{Real Dataset.}
  Using a lensless imaging system to capture images displayed on a screen as measurements is currently the primary method of acquiring datasets in the lensless field.
  We used this approach to collect datasets from various sources, including Human3.6M, MPII, COCO, 3DPW, and MIP-INF-3DHP datasets, which we refer to as  LenslessHuman3.6M, LenslessMPII, LenslessMIP-INF-3DHP, LenslessCOCO, and Lensless3DPW, respectively. 
  We also reclassified the training dataset and validation dataset on LenslessHuman3.6M, referred to as train-LenslessHuman3.6M and eval-LenslessHuman3.6M, respectively.
  Moreover, we capture the real scenes of different individuals, backgrounds, and light intensities using our lensless imaging system.

    \item
   \textbf{Simulated Dataset.}
   The mathematical imaging model enables the conversion (simulation) of images captured by RGB cameras into measurements captured by lensless imaging systems. 
   This technique is commonly employed in lensless imaging fields for testing systems and imaging methods. Leveraging this, we converted numerous human pose datasets into simulated lensless measurements. These simulated datasets extended our training set and provided a quick validation of our method.
   
\end{itemize}



\subsection{Implementation Details}
The dimensions of the raw measurements are $1280 \times 1024 \times 3$.
Prior to commencing the experiments, we first crop and resize the lensless measurements. 
These measurements are pre-processed to $224 \times 224 \times 3$ before being fed into the network.
Following this pre-processing step, the lensless measurements inputted into the global perceptual layer are $224 \times 224$, generating spatial features at resolutions of $(7 \times 7, 14 \times 14, 28 \times 28,$ and $56 \times 56)$.
During the generation of mesh-aligned features, the SMPL mesh is down-sampled using a pre-computed downsampling matrix provided in ~\cite{kolotouros2019convolutional}, resulting in a reduction of vertex count from 6890 to 431.
The regressors $\mathcal{R}_t$ follow a congruent architectural framework to the regressor present in HMR~\cite{kanazawaHMR18}, albeit with slight variations in their input dimensions. 
Our network is trained using the Adam optimizer~\cite{kingma2014adam} with a learning rate of $5 \times 10^{-5}$ and a batch size of 80 on 4 NVIDIA RTX2080 Ti GPUs. No learning rate decay is applied during training.

\begin{figure}[h]
  \centering
   \includegraphics[width=0.95\linewidth]{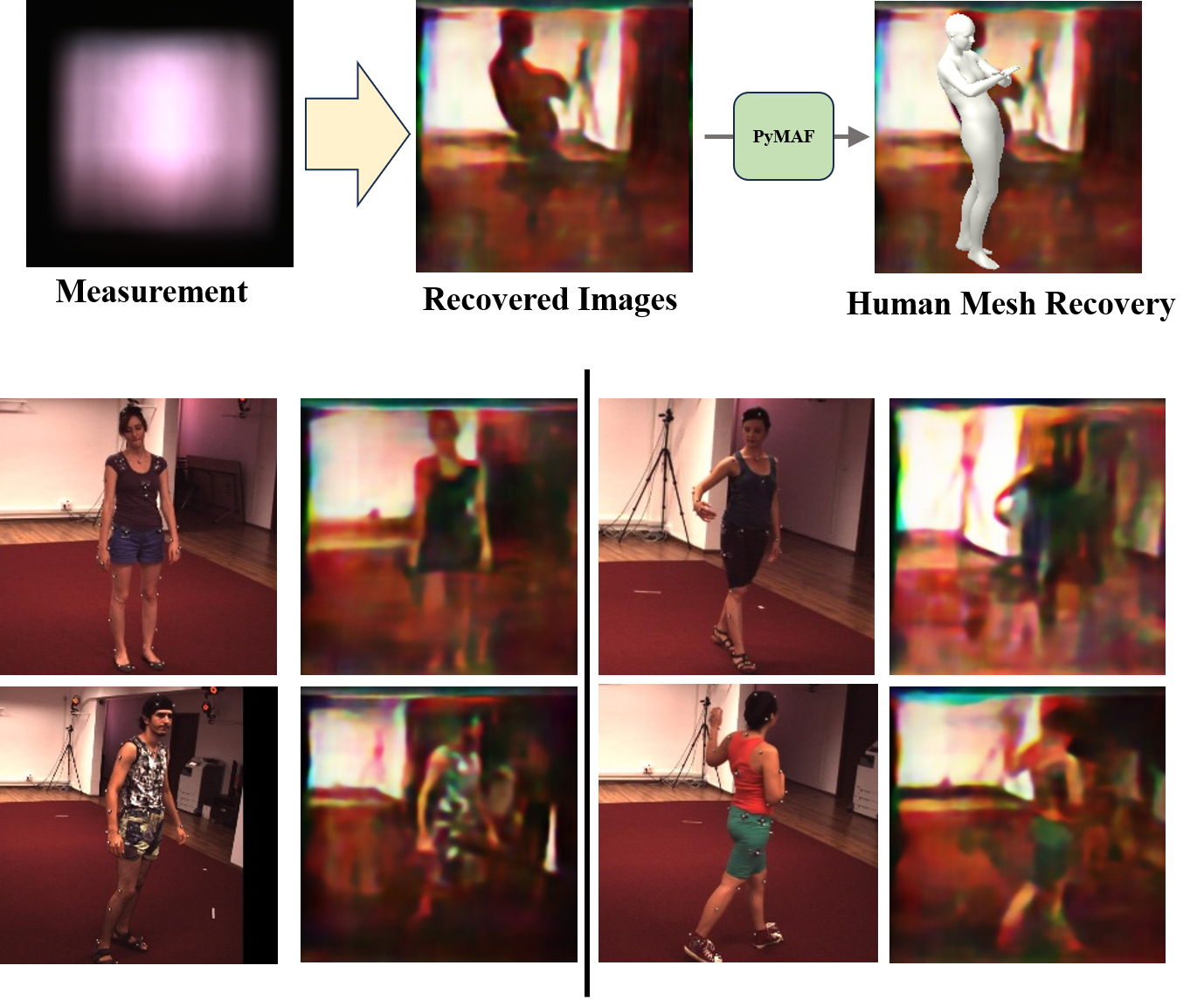}

   \caption{At the top is the frame of our baseline. Lensless measurements first recover the image by reconstruction methods and then estimate human pose and shape by PyMAF.
   The bottom image compares the recovered image with the original image, and we can see that the quality of the recovered image has dropped dramatically.
   }
   \label{fig:4.4}
\end{figure}

\subsection{3D Human Pose and Shape Comparison.}

\begin{figure*}[t]
  \centering
   \includegraphics[width=0.95\linewidth]{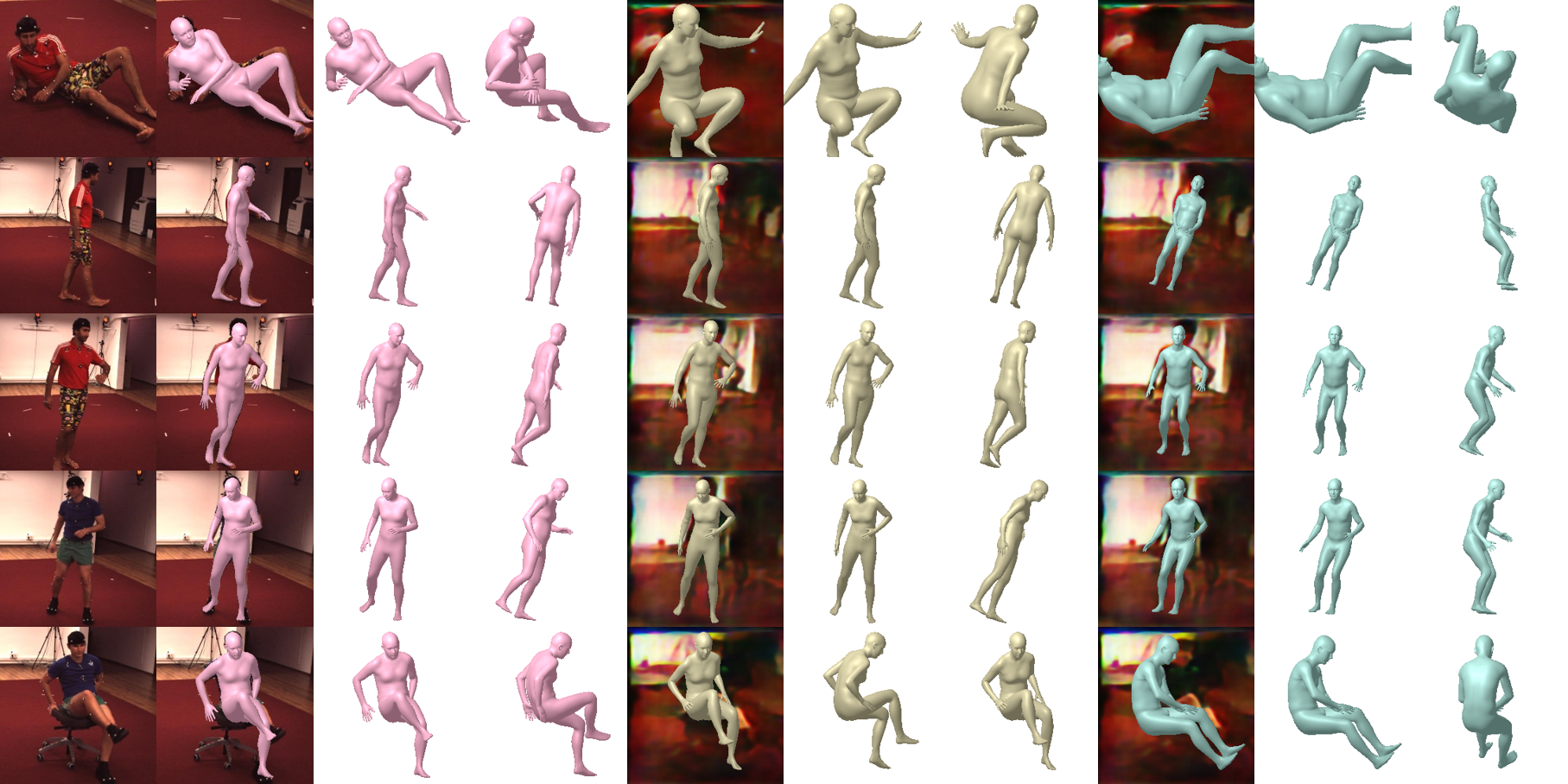}

   \caption{\textbf{Qualitative results of LPSNet  on challenging LenslessHuman3.6M.} The results on the left are from our LPSNet estimation, those in the center are from our baseline (PyMAF) output, and those on the right are from the fine-tuned baseline (PyMAF$^\dagger$) output. LPSNet can be seen to be better than the baselines.}
   \label{fig:4.5}
\end{figure*}

\textbf{Baseline Approach.}
We are the first work to perform human pose and shape estimation through lensless measurements. Given the absence of other valid methods for comparison, we designed a baseline approach that approaches this as a two-stage task. The baseline architecture is depicted in ~\cref{fig:4.4}.
In the first stage, we utilized the lensless image reconstruction method Rego \textit{et. al.}~\cite{rego2021robust}, chosen for its wide applicability in recovering images acquired by lensless imaging systems. In the second stage, PyMAF is used to estimate human pose and shape.

\begin{table}[h]
  \centering
  \small
  \begin{tabular}{c|ccc}
    \toprule
    Method & MPJPE $\downarrow$ & PA-MPJPE$\downarrow$ & PVE $\downarrow$ \\
    \midrule
    baseline (PyMAF) & 257.07 & 121.73 & 277.18 \\
    baseline (PyMAF$^\dagger$) & 126.28 & \textbf{81.37} & 151.60 \\
    LPSNet & \textbf{119.20} & 81.52 & \textbf{134.74} \\
    \bottomrule
  \end{tabular}
  \caption{Comparison with baseline. The baseline is a combination of the two-stage approach using existing methods. LPSNet outperforms the baseline on the LenslessHuman3.6M datasets.}
  \label{tab:1}

\end{table}

\textbf{Comparison with Baseline.}
We present the results of the quantitative comparative evaluation on the LenslessHuman3.6M dataset in ~\cref{tab:1}. 
Notably, images recovered from lensless measurements exhibit lower quality. Directly comparing the baseline (PyMAF) with our approach is challenging and unfair. Therefore, we fine-tuned PyMAF on the train-LenslessHuman3.6M dataset, denoted as PyMAF$^\dagger$.
Referring to ~\cref{tab:1}, we observe significant improvements in the indicators of baseline(PyMAF$^\dagger$). Specifically, compared with the baseline(PyMAF$^\dagger$), the MPJPE of LPSNet on the LenslessHuman3.6M dataset is reduced by 7.08 mm.

As shown in ~\cref{tab:1}, our LPSNet achieved competitive results compared to the baseline approach (PyMAF$^\dagger$). LPSNet demonstrates more significant improvements in the MPJPE and PVE metrics. However, we would argue that the PA-MPJPE metric may not fully reveal the performance of the mesh-image alignment, as it is calculated as MPJPE after rigid alignment.

Baseline (PyMAF) has difficulty performing accurate pose estimation on low-quality recovered images. However, the outputs of our LPSNet exhibit significant improvements compared to the two baseline methods, especially for limb estimation.
Qualitative comparison results are illustrated in ~\cref{fig:4.5}.

\begin{table}[h]
  \centering
  \small
  
 \begin{adjustbox}{width=\columnwidth}
  \begin{tabular}{c|ccc}
    \toprule
    Method   & MPJPE $\downarrow$ & PA-MPJPE$\downarrow$ & PVE $\downarrow$  \\
    \midrule
    w/o MSFDcoder and DHAS & 142.13 & 92.20 & 161.07\\
    w/o DHAS & 139.34 & 92.56 & 158.97   \\
    w/o DHAS (2D keypoint) & 123.30 & 83.41 & 138.67\\
    w/o DHAS (IUV) & 129.70 & 86.83 & 146.7 \\
    LPSNet & \textbf{119.20} & \textbf{81.52} & \textbf{134.74} \\
    \bottomrule
  \end{tabular}
  \end{adjustbox}
  \caption{Ablation study on LPSNet. DHAS stands for Double-Head Auxiliary Supervision.} 
  \label{tab:3}

\end{table}

\begin{figure*}[h]
  \centering
   \includegraphics[width=0.95\linewidth]{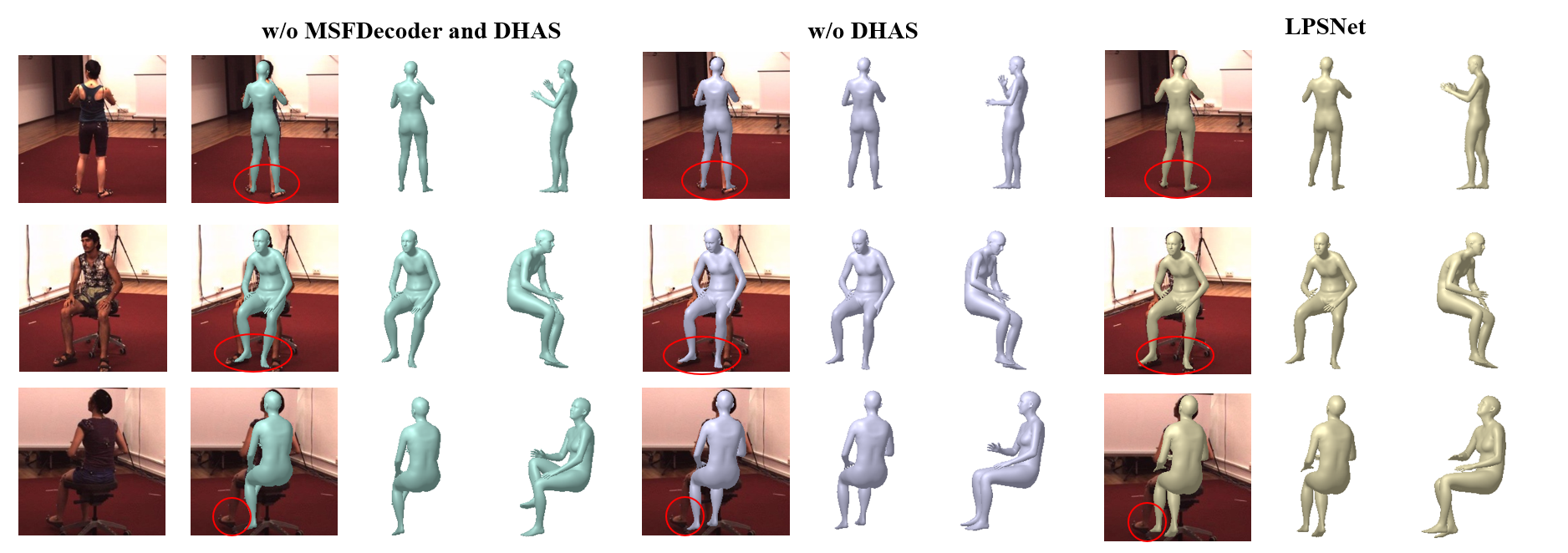}

   \caption{\textbf{Qualitative results of LPSNet on challenging LenslessHuman3.6M.}We can see that the results of LPSNet are better, and the results are improved with the addition of the double-head assisted supervision mechanism.}
   \label{fig:4.7}
\end{figure*}

\subsection{Ablation Study}
In this section, we conduct ablation studies on LenslessHuman3.6M under various settings to validate the effectiveness of the key components proposed in our method. 

All ablation variants were trained and tested on LenslessHuman3.6M, which is derived from the Human3.6M dataset. The Human3.6M dataset includes ground-truth 3D labels and serves as a widely used benchmark in 3D human pose and shape estimation.

\textbf{Multi-Scale Lensless Feature Decoder.}
In LPSNet, the decoder primarily decodes the global information of lensless measurements encoded by the diffuser, which plays a crucial role in our end-to-end system.
Our MSFDecoder is essential for efficient feature extraction from lensless measurements. We have adapted another variant of the decoder to verify this.
Specifically, we simplified the global perception layer of the decoder to a series of convolutional layers and deconvolutional layers.
We use this simpler decoder in the experiment(w/o MSFDcoder and DHAS) instead of MSFDecoder.
As shown in ~\cref{tab:3}, comparing the result of experiments(w/o MSFDcoder and DHAS) and experiment(w/o DHAS), we can see that MSFDecoder improves the accuracy of the human pose and shape estimation significantly.
Through the qualitative experiment in ~\cref{fig:4.7}, we can also see that the human pose and shape estimation has a better alignment after using the MSFDecoder.
Note that neither experiment(w/o MSFDcoder and DHAS) nor the experiment(w/o DHAS) used auxiliary supervision.

\textbf{Double-Head Auxiliary Supervision.}
Double-Head Auxiliary Supervision is mainly used to improve the accuracy of human limbs.
For each auxiliary supervisory head, we conducted ablation experiments respectively. 
Referring to the result in ~\cref{tab:3}, LPSNet has significant improvements in the MPJPE and the PVE metrics.
~\cref{fig:4.7} shows additional qualitative comparison results. We can see that the alignment of the human pose and shape is better with the addition of the Double-Head Auxiliary Supervision mechanism.




\begin{figure}[h]
    \small
    \begin{minipage}[t]{0.26\textwidth}
        \centering
        \includegraphics[width=\textwidth]{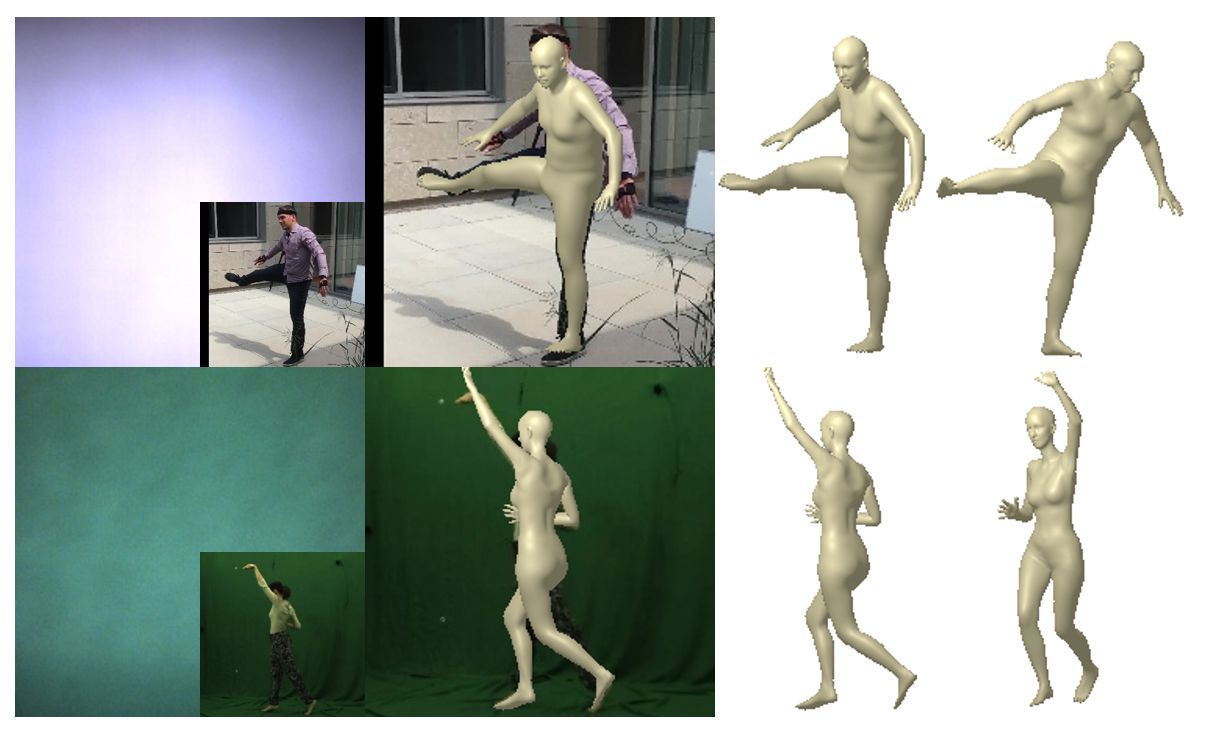}
        \centerline{(a) Result on more datasets}
    \end{minipage}%
    \begin{minipage}[t]{0.22\textwidth}
        \centering
        
        \includegraphics[width=\textwidth]{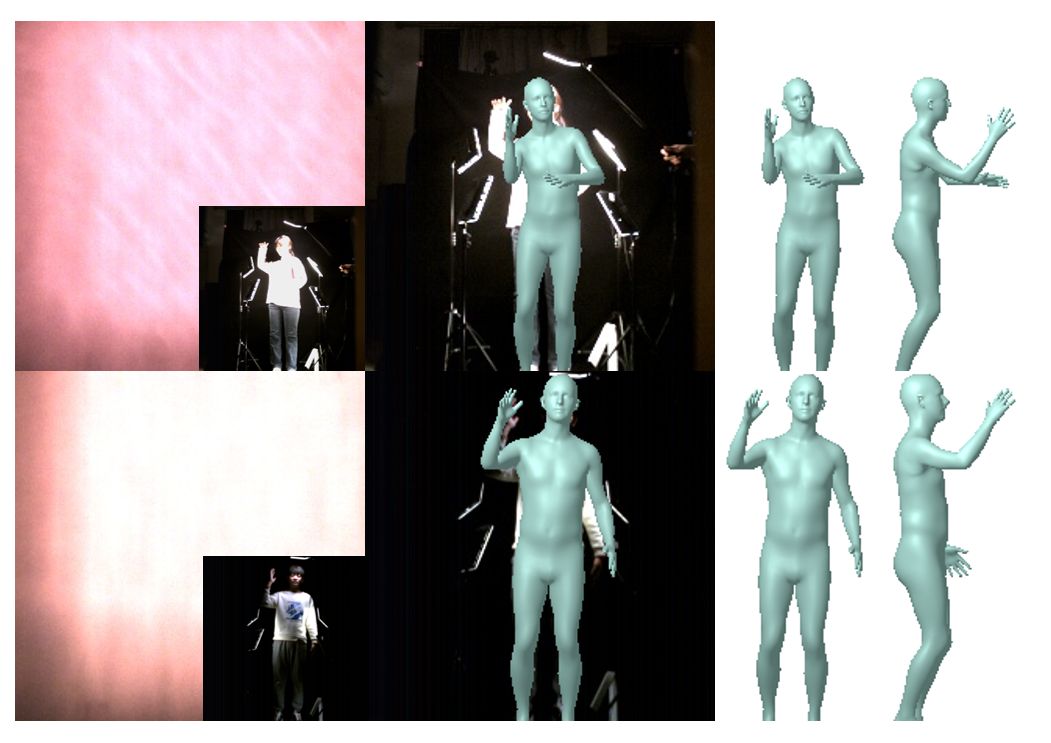}
        \centerline{(b) Result on real scenes}
    \end{minipage}
    \caption{Results on more datasets and real scenes. From left to right for each set of figures: lensless measurement, alignment of results with RGB images, and 3D results shown in different views.}
    \vspace {-0.3cm} 
    \label{fig:4.5.1}
\end{figure}
\subsection{Results on more Datasets and Real Scene}
~\cref{fig:4.5.1} (a) shows the qualitative results on various datasets after using the mixed datasets for training.
Specifically, we collected the MPII, COCO, 3DPW, MIP-INF-3DHP, and LSP datasets through our lensless imaging system and mixed them for training and testing.

In addition, we also capture the real scenes of two different people, in different backgrounds and different light intensities using our lensless imaging system. ~\cref{fig:4.5.1} (b) illustrates the experimental results of our method on these real scenes, indicating its suitability for real-world applications.

\section{Conclusion and Discussion}

\textbf{Conclusion.}
In this paper, we propose the first end-to-end framework to recover 3D human poses and shapes from lensless measurements to the best of our knowledge.
Specifically, we design a multi-scale lensless feature decoder, which can effectively decode the information produced by the lensless imaging system. 
We also propose a double auxiliary supervision mechanism to improve the accuracy of human limb end estimation.
Experimental results show that our method can achieve end-to-end human pose and shape estimation through lensless measurements.

\textbf{Limitations and Future Work.}
Our work is an initial stride step in this direction, however, the results are still not as robust as those achieved by traditional methods for estimating the human body from RGB images. 
The current scarcity of lensless datasets hampers our ability to pre-train the MSFDecoder using a large dataset like ImageNet~\cite{5206848}, thereby limiting the generalizability of our approach. 
We will continue this work in the future, focusing on generating more extensive lensless datasets and progressing towards practical applications.

\textbf{Acknowledgments.}
This work was supported in part by the National Key R\&D Program of China (2023YFC3082100), the National Natural Science Foundation of China (62122058 and 62171317), and the Science Fund for Distinguished Young Scholars of Tianjin (No. 22JCJQJC00040).

{
    \small
    \bibliographystyle{ieeenat_fullname}
    \bibliography{main}

\begin{thebibliography}{38}
\providecommand{\natexlab}[1]{#1}
\providecommand{\url}[1]{\texttt{#1}}
\expandafter\ifx\csname urlstyle\endcsname\relax
  \providecommand{\doi}[1]{doi: #1}\else
  \providecommand{\doi}{doi: \begingroup \urlstyle{rm}\Url}\fi

\bibitem[Adams et~al.(2017)Adams, Boominathan, Avants, Vercosa, Ye, Baraniuk, Robinson, and Veeraraghavan]{adams2017single}
Jesse~K Adams, Vivek Boominathan, Benjamin~W Avants, Daniel~G Vercosa, Fan Ye, Richard~G Baraniuk, Jacob~T Robinson, and Ashok Veeraraghavan.
\newblock {Single-Frame 3D Fluorescence Microscopy with Ultraminiature Lensless FlatScope}.
\newblock \emph{Sci Adv.}, 3\penalty0 (12):\penalty0 e1701548, 2017.

\bibitem[Antipa et~al.(2018)Antipa, Kuo, Heckel, Mildenhall, Bostan, Ng, and Waller]{antipa2018diffusercam}
Nick Antipa, Grace Kuo, Reinhard Heckel, Ben Mildenhall, Emrah Bostan, Ren Ng, and Laura Waller.
\newblock {DiffuserCam: Lensless Single-exposure 3D Imaging}.
\newblock \emph{Optica}, 5\penalty0 (1):\penalty0 1--9, 2018.

\bibitem[Arnab et~al.(2019)Arnab, Doersch, and Zisserman]{arnab2019exploiting}
Anurag Arnab, Carl Doersch, and Andrew Zisserman.
\newblock {Exploiting Temporal Context for 3D Human Pose Estimation in the ild}.
\newblock In \emph{IEEE Conf. Comput. Vis. Pattern Recog.}, pages 3395--3404, 2019.

\bibitem[Asif et~al.(2016)Asif, Ayremlou, Sankaranarayanan, Veeraraghavan, and Baraniuk]{asif2016flatcam}
M~Salman Asif, Ali Ayremlou, Aswin Sankaranarayanan, Ashok Veeraraghavan, and Richard~G Baraniuk.
\newblock {Flatcam: Thin, Lensless Cameras using Coded Aperture and Computation}.
\newblock \emph{IEEE Trans Comput Imaging.}, 3\penalty0 (3):\penalty0 384--397, 2016.

\bibitem[Bogo et~al.(2016)Bogo, Kanazawa, Lassner, Gehler, Romero, and Black]{Bogo:ECCV:2016}
Federica Bogo, Angjoo Kanazawa, Christoph Lassner, Peter Gehler, Javier Romero, and Michael~J. Black.
\newblock {Keep it {SMPL}: Automatic Estimation of {3D} Human Pose and Shape from a Single Image}.
\newblock In \emph{{Eur. Conf. Comput. Vis.}} Springer International Publishing, 2016.

\bibitem[Boominathan et~al.(2016)Boominathan, Adams, Asif, Avants, Robinson, Baraniuk, Sankaranarayanan, and Veeraraghavan]{boominathan2016lensless}
Vivek Boominathan, Jesse~K Adams, M~Salman Asif, Benjamin~W Avants, Jacob~T Robinson, Richard~G Baraniuk, Aswin~C Sankaranarayanan, and Ashok Veeraraghavan.
\newblock {Lensless Imaging: A Computational Renaissance}.
\newblock \emph{IEEE Signal Process Mag.}, 33\penalty0 (5):\penalty0 23--35, 2016.

\bibitem[Boominathan et~al.(2020)Boominathan, Adams, Robinson, and Veeraraghavan]{boominathan2020phlatcam}
Vivek Boominathan, Jesse~K Adams, Jacob~T Robinson, and Ashok Veeraraghavan.
\newblock {Phlatcam: Designed Phase-Mask Based Thin Lensless Camera}.
\newblock \emph{IEEE Trans. Pattern Anal. Mach. Intell.}, 42\penalty0 (7):\penalty0 1618--1629, 2020.

\bibitem[Cho et~al.(2022)Cho, Youwang, and Oh]{cho2022cross}
Junhyeong Cho, Kim Youwang, and Tae-Hyun Oh.
\newblock Cross-attention of disentangled modalities for 3d human mesh recovery with transformers.
\newblock In \emph{Eur. Conf. Comput. Vis.}, pages 342--359. Springer, 2022.

\bibitem[Deng et~al.()Deng, Dong, Socher, Li, Li, and Fei-Fei]{5206848}
J. Deng, W. Dong, R. Socher, L. Li, Kai Li, and Li Fei-Fei.
\newblock Imagenet: A large-scale hierarchical image database.
\newblock In \emph{IEEE Conf. Comput. Vis. Pattern Recog. Workshops (CVPR Workshops)}.

\bibitem[Georgakis et~al.(2020)Georgakis, Li, Karanam, Chen, Ko{\v{s}}eck{\'a}, and Wu]{georgakis2020hierarchical}
Georgios Georgakis, Ren Li, Srikrishna Karanam, Terrence Chen, Jana Ko{\v{s}}eck{\'a}, and Ziyan Wu.
\newblock {Hierarchical Kinematic Human Mesh Recovery}.
\newblock In \emph{Eur. Conf. Comput. Vis.}, pages 768--784. Springer, 2020.

\bibitem[Goel et~al.(2023)Goel, Pavlakos, Rajasegaran, Kanazawa, and Malik]{Goel_2023_ICCV}
Shubham Goel, Georgios Pavlakos, Jathushan Rajasegaran, Angjoo Kanazawa, and Jitendra Malik.
\newblock {Humans in 4D: Reconstructing and Tracking Humans with Transformers}.
\newblock In \emph{{Int. Conf. Comput. Vis.}}, pages 14783--14794, 2023.

\bibitem[Goodman(2005)]{goodman2005introduction}
Joseph~W Goodman.
\newblock \emph{Introduction to Fourier optics}.
\newblock Roberts and Company publishers, 2005.

\bibitem[G{\"u}ler et~al.(2018)G{\"u}ler, Neverova, and Kokkinos]{guler2018densepose}
R{\i}za~Alp G{\"u}ler, Natalia Neverova, and Iasonas Kokkinos.
\newblock {Densepose: Dense Human Pose Estimation in the Wild}.
\newblock In \emph{Int. Conf. Comput. Vis.}, pages 7297--7306, 2018.

\bibitem[Ionescu et~al.(2013)Ionescu, Papava, Olaru, and Sminchisescu]{ionescu2013human3}
Catalin Ionescu, Dragos Papava, Vlad Olaru, and Cristian Sminchisescu.
\newblock {Human3. 6M: Large Scale Datasets and Predictive Methods for 3D Human Sensing in Natural Environments}.
\newblock \emph{IEEE Trans. Pattern Anal. Mach. Intell.}, 36\penalty0 (7):\penalty0 1325--1339, 2013.

\bibitem[Joo et~al.(2021)Joo, Neverova, and Vedaldi]{joo2021exemplar}
Hanbyul Joo, Natalia Neverova, and Andrea Vedaldi.
\newblock {Exemplar Fine-Tuning for 3D Human Model Fitting Towards In-the-Wild 3D Human Pose Estimation}.
\newblock In \emph{Int. Conf. 3D. Vis.}, pages 42--52. IEEE, 2021.

\bibitem[Kanazawa et~al.(2018)Kanazawa, Black, Jacobs, and Malik]{kanazawaHMR18}
Angjoo Kanazawa, Michael~J. Black, David~W. Jacobs, and Jitendra Malik.
\newblock {End-to-end Recovery of Human Shape and Pose}.
\newblock In \emph{IEEE Conf. Comput. Vis. Pattern Recog.}, 2018.

\bibitem[Kingma and Ba(2014)]{kingma2014adam}
Diederik~P Kingma and Jimmy Ba.
\newblock Adam: A method for stochastic optimization.
\newblock \emph{arXiv preprint arXiv:1412.6980}, 2014.

\bibitem[Kocabas et~al.(2021)Kocabas, Huang, Hilliges, and Black]{kocabas2021pare}
Muhammed Kocabas, Chun-Hao~P Huang, Otmar Hilliges, and Michael~J Black.
\newblock {PARE: Part Attention Regressor for 3D Human Body Estimation}.
\newblock In \emph{Int. Conf. Comput. Vis.}, pages 11127--11137, 2021.

\bibitem[Kolotouros et~al.(2019{\natexlab{a}})Kolotouros, Pavlakos, Black, and Daniilidis]{kolotouros2019learning}
Nikos Kolotouros, Georgios Pavlakos, Michael~J Black, and Kostas Daniilidis.
\newblock {Learning to Reconstruct 3D Human Pose and Shape Via Model-fitting in the Loop}.
\newblock In \emph{Int. Conf. Comput. Vis.}, pages 2252--2261, 2019{\natexlab{a}}.

\bibitem[Kolotouros et~al.(2019{\natexlab{b}})Kolotouros, Pavlakos, and Daniilidis]{kolotouros2019convolutional}
Nikos Kolotouros, Georgios Pavlakos, and Kostas Daniilidis.
\newblock {Convolutional Mesh Regression for Single-Image Human Shape Reconstruction}.
\newblock In \emph{IEEE Conf. Comput. Vis. Pattern Recog.}, pages 4501--4510, 2019{\natexlab{b}}.

\bibitem[Kuo et~al.(2017)Kuo, Antipa, Ng, and Waller]{kuo2017diffusercam}
Grace Kuo, Nick Antipa, Ren Ng, and Laura Waller.
\newblock {DiffuserCam: Diffuser-Based Lensless Cameras}.
\newblock In \emph{Computational Optical Sensing and Imaging}, pages CTu3B--2. Optica Publishing Group, 2017.

\bibitem[Lassner et~al.(2017)Lassner, Romero, Kiefel, Bogo, Black, and Gehler]{Lassner:UP:2017}
Christoph Lassner, Javier Romero, Martin Kiefel, Federica Bogo, Michael~J. Black, and Peter~V. Gehler.
\newblock {Unite the People: Closing the Loop Between 3D and 2D Human Representations}.
\newblock In \emph{IEEE Conf. Comput. Vis. Pattern Recog.}, 2017.

\bibitem[Leroy et~al.(2020)Leroy, Weinzaepfel, Br{\'e}gier, Combaluzier, and Rogez]{leroy2020smply}
Vincent Leroy, Philippe Weinzaepfel, Romain Br{\'e}gier, Hadrien Combaluzier, and Gr{\'e}gory Rogez.
\newblock {SMPLy Benchmarking 3D Human Pose Estimation in the wild}.
\newblock In \emph{Int. Conf. 3D. Vis.}, pages 301--310. IEEE, 2020.

\bibitem[Li et~al.(2022)Li, Yang, Liu, Zhang, Wang, Wang, Yang, and Xia]{li2022simcc}
Yanjie Li, Sen Yang, Peidong Liu, Shoukui Zhang, Yunxiao Wang, Zhicheng Wang, Wankou Yang, and Shu-Tao Xia.
\newblock {SimCC: A Simple Coordinate Classification Perspective for Human Pose Estimation}.
\newblock In \emph{Eur. Conf. Comput. Vis.}, pages 89--106. Springer, 2022.

\bibitem[Lin et~al.(2021{\natexlab{a}})Lin, Wang, and Liu]{lin2021end}
Kevin Lin, Lijuan Wang, and Zicheng Liu.
\newblock End-to-end human pose and mesh reconstruction with transformers.
\newblock In \emph{IEEE Conf. Comput. Vis. Pattern Recog.}, pages 1954--1963, 2021{\natexlab{a}}.

\bibitem[Lin et~al.(2021{\natexlab{b}})Lin, Wang, and Liu]{lin2021mesh}
Kevin Lin, Lijuan Wang, and Zicheng Liu.
\newblock {Mesh Graphormer}.
\newblock In \emph{Int. Conf. Comput. Vis.}, pages 12939--12948, 2021{\natexlab{b}}.

\bibitem[Ozcan and McLeod(2016)]{ozcan2016lensless}
Aydogan Ozcan and Euan McLeod.
\newblock {Lensless Imaging and Sensing}.
\newblock \emph{Annual review of biomedical engineering}, 18:\penalty0 77--102, 2016.

\bibitem[Pavlakos et~al.(2019)Pavlakos, Choutas, Ghorbani, Bolkart, Osman, Tzionas, and Black]{SMPL-X:2019}
Georgios Pavlakos, Vasileios Choutas, Nima Ghorbani, Timo Bolkart, Ahmed A.~A. Osman, Dimitrios Tzionas, and Michael~J. Black.
\newblock {Expressive Body Capture: 3D Hands, Face, and Body from a Single Image}.
\newblock In \emph{IEEE Conf. Comput. Vis. Pattern Recog.}, 2019.

\bibitem[Pavlakos et~al.(2022)Pavlakos, Malik, and Kanazawa]{pavlakos2022human}
Georgios Pavlakos, Jitendra Malik, and Angjoo Kanazawa.
\newblock {Human Mesh Recovery from Multiple Shots}.
\newblock In \emph{IEEE Conf. Comput. Vis. Pattern Recog.}, pages 1485--1495, 2022.

\bibitem[Rego et~al.(2021)Rego, Kulkarni, and Jayasuriya]{rego2021robust}
Joshua~D Rego, Karthik Kulkarni, and Suren Jayasuriya.
\newblock {Robust Lensless Image Reconstruction Via PSF Estimation}.
\newblock In \emph{Proceedings of the IEEE/CVF Winter Conference on Applications of Computer Vision}, pages 403--412, 2021.

\bibitem[Rempe et~al.(2021)Rempe, Birdal, Hertzmann, Yang, Sridhar, and Guibas]{rempe2021humor}
Davis Rempe, Tolga Birdal, Aaron Hertzmann, Jimei Yang, Srinath Sridhar, and Leonidas~J. Guibas.
\newblock {HuMoR: 3D Human Motion Model for Robust Pose Estimation}.
\newblock In \emph{Int. Conf. Comput. Vis.}, 2021.

\bibitem[Stork and Gill(2013)]{stork2013lensless}
David~G Stork and Patrick~R Gill.
\newblock {Lensless Ultra-miniature CMOS Computational Imagers and Sensors}.
\newblock \emph{Proc. Sensorcomm}, pages 186--190, 2013.

\bibitem[Stork and Gill(2014)]{stork2014optical}
David~G Stork and Patrick~R Gill.
\newblock Optical, mathematical, and computational foundations of lensless ultra-miniature diffractive imagers and sensors.
\newblock \emph{International Journal on Advances in Systems and Measurements}, 7\penalty0 (3):\penalty0 4, 2014.

\bibitem[Sun et~al.(2019)Sun, Xiao, Liu, and Wang]{sun2019deep}
Ke Sun, Bin Xiao, Dong Liu, and Jingdong Wang.
\newblock {Deep High-resolution Representation Learning for Human Pose Estimation}.
\newblock In \emph{IEEE Conf. Comput. Vis. Pattern Recog.}, pages 5693--5703, 2019.

\bibitem[Tiwari et~al.(2022)Tiwari, Antic, Lenssen, Sarafianos, Tung, and Pons-Moll]{tiwari22posendf}
Garvita Tiwari, Dimitrije Antic, Jan~Eric Lenssen, Nikolaos Sarafianos, Tony Tung, and Gerard Pons-Moll.
\newblock {Pose-NDF: Modeling Human Pose Manifolds with Neural Distance Fields}.
\newblock In \emph{Eur. Conf. Comput. Vis.}, 2022.

\bibitem[Zhang et~al.(2020)Zhang, Cao, Lu, Ouyang, and Sun]{zhang2020learning}
Hongwen Zhang, Jie Cao, Guo Lu, Wanli Ouyang, and Zhenan Sun.
\newblock {Learning 3D Human Shape and Pose from Dense Body Parts}.
\newblock \emph{IEEE Trans. Pattern Anal. Mach. Intell.}, 44\penalty0 (5):\penalty0 2610--2627, 2020.

\bibitem[Zhang et~al.(2021)Zhang, Tian, Zhou, Ouyang, Liu, Wang, and Sun]{zhang2021pymaf}
Hongwen Zhang, Yating Tian, Xinchi Zhou, Wanli Ouyang, Yebin Liu, Limin Wang, and Zhenan Sun.
\newblock Pymaf: 3d human pose and shape regression with pyramidal mesh alignment feedback loop.
\newblock In \emph{Int. Conf. Comput. Vis.}, pages 11446--11456, 2021.

\bibitem[Zhang et~al.(2019)Zhang, Felsen, Kanazawa, and Malik]{zhang2019predicting}
Jason~Y Zhang, Panna Felsen, Angjoo Kanazawa, and Jitendra Malik.
\newblock {Predicting 3D Human Dynamics from Video}.
\newblock In \emph{Int. Conf. Comput. Vis.}, pages 7114--7123, 2019.

\end{thebibliography}


\begin{thebibliography}{3}
\providecommand{\natexlab}[1]{#1}
\providecommand{\url}[1]{\texttt{#1}}
\expandafter\ifx\csname urlstyle\endcsname\relax
  \providecommand{\doi}[1]{doi: #1}\else
  \providecommand{\doi}{doi: \begingroup \urlstyle{rm}\Url}\fi

\bibitem[Boominathan et~al.(2020)Boominathan, Adams, Robinson, and Veeraraghavan]{boominathan2020phlatcam}
Vivek Boominathan, Jesse~K Adams, Jacob~T Robinson, and Ashok Veeraraghavan.
\newblock {Phlatcam: Designed Phase-Mask Based Thin Lensless Camera}.
\newblock \emph{IEEE Trans. Pattern Anal. Mach. Intell.}, 42\penalty0 (7):\penalty0 1618--1629, 2020.

\bibitem[Goodman(2005)]{goodman2005introduction}
Joseph~W Goodman.
\newblock \emph{Introduction to Fourier optics}.
\newblock Roberts and Company publishers, 2005.

\bibitem[Ionescu et~al.(2013)Ionescu, Papava, Olaru, and Sminchisescu]{ionescu2013human3}
Catalin Ionescu, Dragos Papava, Vlad Olaru, and Cristian Sminchisescu.
\newblock {Human3. 6M: Large Scale Datasets and Predictive Methods for 3D Human Sensing in Natural Environments}.
\newblock \emph{IEEE Trans. Pattern Anal. Mach. Intell.}, 36\penalty0 (7):\penalty0 1325--1339, 2013.

\end{thebibliography}
}


\end{document}